\documentclass[10pt,twocolumn,letterpaper]{article}

\usepackage{iccv}              % To produce the CAMERA-READY version
\usepackage{xcolor}
\usepackage{soul}

\definecolor{darkgreen}{rgb}{0.0, 0.6, 0.0}
\usepackage{multirow}
\definecolor{iccvblue}{rgb}{0.21,0.49,0.74}
\usepackage[pagebackref,breaklinks,colorlinks,allcolors=iccvblue]{hyperref}
\usepackage{tcolorbox}

\def\paperID{12086} % *** Enter the Paper ID here
\def\confName{ICCV}
\def\confYear{2025}

\title{
AuthGuard: Generalizable Deepfake Detection via Language Guidance
}

\author{Guangyu Shen\textsuperscript{1}\thanks{Work done during internship at AWS AI Labs.}\hspace{0.3em}, Zhihua Li\textsuperscript{2}, Xiang Xu\textsuperscript{2}, Tianchen Zhao\textsuperscript{2}, Zheng Zhang\textsuperscript{2}, Dongsheng An\textsuperscript{2},\\Zhuowen Tu\textsuperscript{2}, Yifan Xing\textsuperscript{2}, Qin Zhang\textsuperscript{2}\\\textsuperscript{1}Purdue University \quad 
\textsuperscript{2}AWS AI Labs \\[0.5ex]
}

\newcommand{\xiangx}[1]{\textcolor{red}{#1}}

\usepackage{comment}

\begin{document}
\maketitle
\begin{abstract}
%\yifan{It still reads that we are adding deepfake detection into a MLLM, the motivation of which is not clear. - why does a MLLM need deepfake detection? what's the use-case that it unlocks? I remember R2 of CVPR was confused by this too.} 

Existing deepfake detection techniques struggle to keep-up with the ever-evolving novel, unseen forgeries methods. This limitation stems from their reliance on statistical artifacts learned during training, which are often tied to specific generation processes that may not be representative of samples from new, unseen deepfake generation methods encountered at test time. We propose that incorporating language guidance can improve deepfake detection generalization by integrating human-like commonsense reasoning -- such as recognizing logical inconsistencies and perceptual anomalies -- alongside statistical cues. To achieve this, we train an expert deepfake vision encoder by combining discriminative classification with image-text contrastive learning, where the text is generated by generalist MLLMs using few-shot prompting. This allows the encoder to extract both language-describable, commonsense deepfake artifacts and statistical forgery artifacts from pixel-level distributions. To further enhance robustness, we integrate data uncertainty learning into vision-language contrastive learning, mitigating noise in image-text supervision. Our expert vision encoder seamlessly interfaces with an LLM, further enabling more generalized and interpretable deepfake detection while also boosting accuracy. The resulting framework, \textbf{AuthGuard}, achieves state-of-the-art deepfake detection accuracy in both in-distribution and out-of-distribution settings, achieving AUC gains of 6.15\% on the DFDC dataset and 16.68\% on the DF40 dataset. Additionally, \textbf{AuthGuard} significantly enhances deepfake reasoning, improving performance by 24.69\% on the DDVQA dataset.

\end{abstract}
\section{Introduction}

\begin{figure}
    \centering
\includegraphics[width=\linewidth]{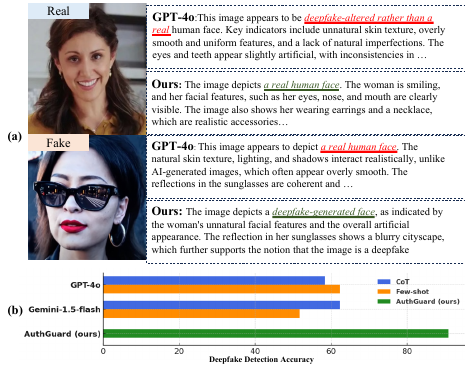}
    \caption{(a) Commercial models perform significantly worse than AuthGuard. For instance, GPT-4o frequently misclassifies low-resolution faces as fake, highlighting its limitations in distinguishing real from fake within general-purpose MLLMs. This underscores the need for a more accurate and lightweight deepfake-specific detection and reasoning model. (b) We leverage the VIC \cite{luo2025ursa} to generate reasoning prompts for GPT-4o and Gemini-1.5 in real/fake classification. Few-shot prompting includes four labeled images (two real, two fake) for context. AuthGuard significantly outperforms commercial models on the DD-VQA \cite{zhang2024common} dataset.
}
    \label{fig:first}
\end{figure}

Recent advances in generative AI~\cite{ruiz2023dreambooth,rombach2022high,ho2020denoising,goodfellow2020generative} have now enabled the creation of hyper-realistic facial manipulations, increasingly blurring the line between real and synthetic imagery. This raises significant risks, including misinformation, identity fraud, and the erosion of public trust in visual evidence. Imagine a breaking news featuring a world leader announcing a sudden policy change, sparking panic in financial markets -- only to later be revealed as a deepfake fabricated to manipulate public opinion. Or picture a virtual job interview with a recruiter, but the person behind the screen is a deepfake avatar orchestrating a phishing scam. As digital media increasingly shapes both public discourse and personal interactions, the ability to detect and analyze deepfake images is becoming more and more essential for safeguarding authenticity~\cite{xu2024principles}.

However, existing deepfake detection methods -- such as~\cite{sbi, ucf, lsda, x-ray, zhao2021learning, spsl, srm} -- struggle to keep pace with the rapid advancements in generative AI. Most approaches focus either on enhancing the vision encoders \cite{sbi,ucf,lsda}  or employing data-driven classification models trained on exemplar attack samples~\cite{nguyen2024laa, xu2023tall, chen2022self}. While these methods are effective against known manipulations, they often fail catastrophically when confronted with novel, unseen deepfakes, leaving critical vulnerabilities in real-world applications as new forgery methods rapidly emerge. This limitation arises because these methods rely heavily on recognizing \textbf{\textit{statistical deepfake artifacts}} tied to specific generation processes -- patterns that exist within their labeled training data~\cite{mit_detect_fakes,abdullah2024analysis,pei2024deepfake} -- which may not appear in the test data. In contrast, humans approach deepfake detection fundamentally differently: instead of relying on statistical cues, they apply commonsense reasoning and describe inconsistencies using natural language~\cite{groh2022deepfake}. As also discussed in \cite{korshunov2020deepfake,groh2022deepfake}, many cross-domain deepfake detection errors align with language describable patterns, indicating that existing methods overlook such key semantic insights.

To enhance generalization in deepfake detection, we propose {\textbf{AuthGuard}}, a unified deepfake detection and reasoning framework that captures both statistical deepfake artifacts and commonsense deepfake artifacts -- human-interpretable, language-describable features that are independent of specific generative models. To achieve this, we develop an automatic data generation pipeline, leveraging state-of-the-art public Multimodal Large Language Models (MLLMs). By incorporating real or fake labels as contextual prompts, we generate 114k high-quality image-text pairs, with each text explaining why a face image appears fake or real. Using this dataset, we train an expert deepfake vision encoder by image-text combining contrastive learning with standard binary classification. This allows the trained vision encoder to capture broad semantic relationships for better generalization while enabling precise distinctions between real and deepfake images. Meanwhile, we mitigate text noise through probabilistic embedding~\cite{shi2019probabilistic}, ensuring more robust cross-modal feature alignment. To effectively balance the contributions from statistical and commonsense deepfake artifacts, we introduce an adaptive aggregation mechanism which uses a light-weight adaptor network to generate input-dependent probabilistic weights, dynamically combining these artifacts into a single visual representation. The aggregated representation enhances generalization and addresses challenges in adapting to novel attacks. Finally, we integrate the aggregated visual representation into the LLaVA architecture~\cite{llava-1.5} through instruction tuning on our curated dataset, creating a flexible framework where the LLM functions as a plug-and-play component for deepfake reasoning and explainability. 

In summary, we make the following contributions: 
\begin{enumerate}
\item We propose \textbf{AuthGuard}, a unified vision-language model that seamlessly integrates deepfake detection, reasoning, and analysis with enhanced generalization. To the best of our knowledge, we are the first work to achieve such unification in the deepfake domain. 

\item We develop a simple yet effective training strategy for learning an expert deepfake vision encoder that captures both commonsense and statistical deepfake artifacts. This strategy employs a text-regularized representation learning method that uses MLLM-generated text data to reduce overfitting to statistical patterns in the training data, and incorporates a small adaptor to combine contrastive and discriminative features dynamically.

\item Through thorough benchmarking, we demonstrate that AuthGuard outperforms state-of-the-art methods in accuracy, generalization and reasoning. Specifically, AuthGuard achieves a 6.15\% improvement in deepfake detection accuracy on the DFDC dataset \cite{dfdc} in the cross-dataset setting and a 24.36\% increase in reasoning accuracy on the DDVQA dataset \cite{zhang2024common}. 
\end{enumerate}

\section{Related Work}
\begin{comment}
\noindent \textbf{Deepfake Generation}: 
Deepfakes can be categorized into four main categories~\cite{tolosana2020deepfakes, xu2024principles}:
(1) Entire face synthesis: in this category, the entire facial images are synthesized or generated by generative models such as Generative Adversarial Networks (GANs)~\cite{shen2018faceid} or Diffusion models \cite{ye2023ip-adapter}.
(2) Facial identity swap: Face swapping ~\cite{faceswap, zhao2023diffswap} techniques typically involve two identities, aiming to replace the facial identity of one person with a victim's identity.
(3) Facial attribute manipulation: This category encompasses the modification of specific facial attributes~\cite{zhang2018generative}, such as gender, age, or the addition/removal of facial accessories.
(4) Face reenactment: These techniques~\cite{face2face, neuraltexture} modify facial expressions or movements of action units to replace expressions or simulate the speech of another individual.
In this paper, 'deepfake' refers specifically to face forgeries, excluding full-image synthesis.
%With the recent rapid advancements in generative models and hardware capabilities, the generation of deepfakes, even in real-time, has become significantly more accessible.
\end{comment}
\noindent \textbf{Deepfake Detection} 
As deepfake generation technology progresses, datasets now encompass a wider variety of attack types to evaluate the accuracy, robustness, and generalization of detection methods~\cite{ff++, li2020celeb, dfdc, yuan2024unified, yan2024df40}.
Traditionally, deepfake detectors have relied on binary image classifiers \cite{x-ray,zhao2021learning, chen2022self}. Recent advancements aim to enhance both accuracy and generalizability by optimizing various components of these detectors \cite{huang2023implicit,zhao2021learning,nguyen2024laa,lsda}. One approach, as seen in~\cite{zhao2021learning, sbi}, enhances the training data by generating synthetic samples that combine source and target images, thus enriching the dataset and potentially improving detector robustness. Another line of work aims to improve representation learning by adopting more powerful backbones and introducing novel training mechanisms. For instance, ~\cite{chen2022self} introduces additional blending mask learning in the training and ~\cite{ucf} decomposes images to reveal common forgery features, enabling the detector to learn more generalized characteristics of deepfakes.
Similarly, \cite{lsda} demonstrates that using representations from a wider variety of forgeries helps create more generalizable decision boundaries, reducing overfitting to method-specific features.

%The authors of the study cited in \cite{yan2024df40} conducted comprehensive experiments to evaluate the detection performance against a diverse range of deepfake attacks. 
%Their findings indicated that the model employing the CLIP-Large \cite{clip} architecture, trained on image and text pairs from the generic domain, exhibited superior accuracy and generalization capabilities compared to other models evaluated.

%While these innovations have advanced deepfake detection accuracy to surpass human-level performance on seen attack types, the generalizability of these methods remains a challenge when encountering unseen deepfakes types, also known as out-of-distribution samples.
%Furthermore, traditional detectors provide only confidence scores of the predicted labels, indicating whether an image is classified as real or fake, without providing insights into the underlying features or characteristics that led to the classification. Consequently, users may find it challenging to understand the rationale behind the detection decisions, which can limit the practical utility of these systems in nuanced or complex scenarios.

\begin{figure*}[t!]
    \centering
    \includegraphics[width=1.0\textwidth]{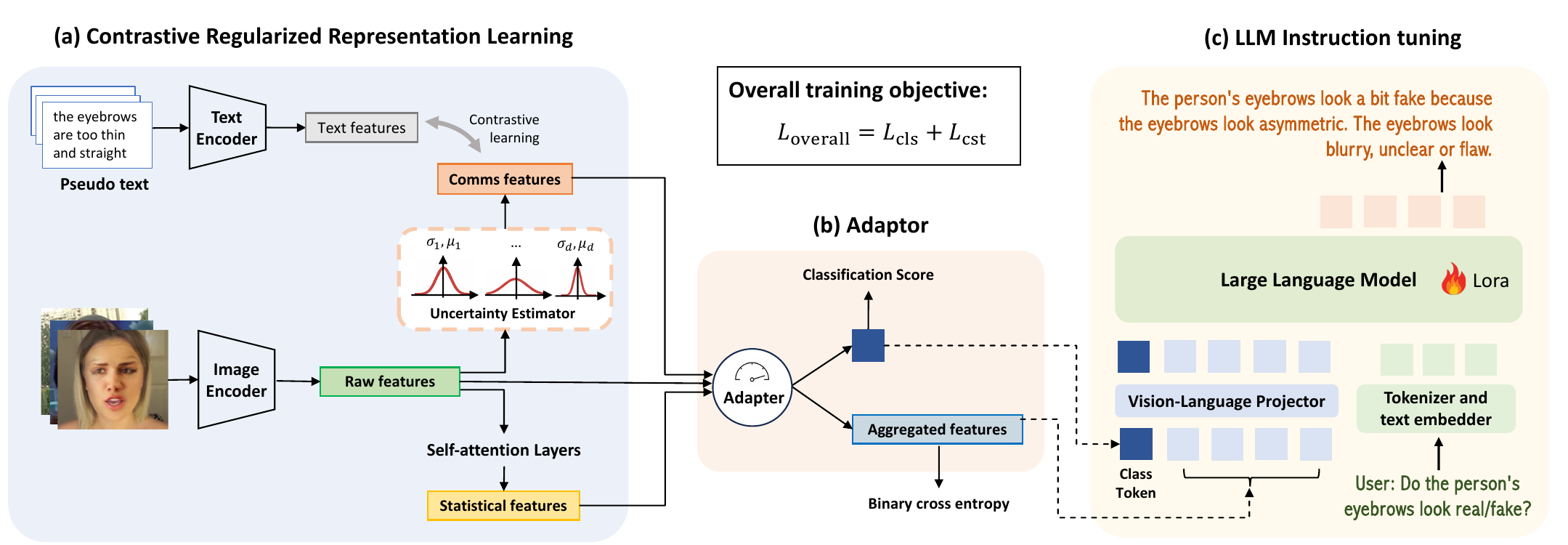}
    \caption{The framework of AuthGuard comprises two main components: (a) and (b) for expert vision representation learning, and (c) for LLM-based deepfake reasoning, where the LLM utilizes image tokens generated from the vision module. In the representation learning module, deepfake artifact learning is divided into commonsense and statistical artifacts. Commonsense artifacts are learned through vision contrastive learning and refined with probabilistic embedding to mitigate label noise. These are then combined with data-driven, forgery-specific artifacts via an adaptive router to extract expert deepfake features. Finally, the adapter’s output class token is concatenated with patch-wise tokens from deeper layers and fed into the LLM for reasoning.}
    %Notably, the LLM reasoning module is plug-and-play; without it, our pipeline functions as a standalone deepfake classification model. %\eric{consider adding information about which part is inference pipeline, training stages 1 and 2, and which part of the pipeline is freezed at given stage of training} \qin{You may want to make it more modular for better readability. Also the text needs to be bigger and use the same font. } \yifan{+1 on both points the left and right is not obvious. Also, here the division mentions ``semantic and distributional'' artifacts whereas in the intro it's semantic and forgery-specific (up to debate if this is the best choice). We should make sure it's consistent across the paper. Also, why is the semantic artifact embedding pointing to the forgery-specific artifact embedding in sequence, shouldn't these two be separate?}
    
    \label{fig:contrastive}
\end{figure*}

\noindent \textbf{Adapting Vision-Language Models for Deepfake Detection} \ 
Recent works~\cite{antifakeprompt, zhang2024common} have explored using instruction tuning of vision-language models~\cite{blip, panagopoulou2023x} to perform deepfake detection by framing it as a visual question answering (VQA) task. One approach~\cite{antifakeprompt} tunes soft prompts to classify images as real or fake, but it only provides a binary yes/no answer, missing the opportunity for more informative, explainable outputs. Another study by Zhang \etal~\cite{zhang2024common} performs annotation on the FaceForensics++ dataset~\cite{ff++} with explanations for fake images and fine-tunes a BLIP model~\cite{blip} to generate explanations to generate reasoning behind why an image appears fake. However, this method suffers from limited annotated data and the high cost of manual labeling. Moreover, it does not account for the critical role of statistical artifacts in deepfake detection, which compromises its robustness in dealing with images containing subtle statistical discrepancies which can not be described by natural language.

%\yifan{we should also highlight the difference in method / design out side of the performance sicne DD-VQA is the closest to ours.} \eric{related work is a bit too long} \yifan{+1, we can take out deepfake generation and much reduce MLLM which are not the focus of our paper.}

%In the following sections, we qualitatively examined the limitations of existing VLMs for the deepfake detection task and analyzed the underlying reasons behind these limitations, which are presented in the Sec.~\ref{sec:evaluation}. We then propose our corresponding solutions to improve detection accuracy, explainability, and generalization capabilities in Sec.~\ref{sec:method}. Our approach involves curating our own image-text pair datasets, employing contrastive learning between image and text modalities, and further integrating with recent language models to generate accurate and fine-grained explanations corresponding to the detection results.

\section{Method}
\label{sec:method}
\cref{fig:contrastive} illustrates the overall framework for \textbf{AuthGuard}. 
We first develop an expert deepfake encoder that captures both statistical and commonsense deepfake features by combining classification with vision-language contrastive learning. Specifically, commonsense deepfake features refer to language-describable and semantically meaningful artifacts learned through pseudo-text pairs, while statistical deepfake features include artifacts that are often imperceptible to humans, such as GAN fingerprints. An adaptor combines these two visual features, creating a balanced representation for more generalized deepfake detection. After the deepfake vision encoder is trained, we integrate it with an LLM, enabling plug-and-play deepfake reasoning and interpretability. In the following sections, we describe our pseudo-text generation process in \cref{sec:datagen}, detail the design and training of the vision encoder in \cref{align} and \cref{uncertainty}, and discuss its integration with an LLM in \cref{sec:llm}.

% \textbf{Architecture} We adopt the architectural design of LLaVA~\cite{llava} to structure our implementation. Specifically
%\xiangx{Lack of transition, To address the limitation/observation, we proposed a two-stage training paradigram harsessing MLLM.....}
%\Zheng{There is no need to detail the entire process since we already have it in introduction and figure 1 caption. Just tell readers each 3.x is for which} \eric{agree, current paragraph should be put later into a subsection named "inference". Here you should put a high level summary of intuition/motivation of your method.}
%For a given input image, the vision encoder extracts a sequence of vision tokens and the last layer class embeddings. These vision tokens are concatenated and then projected into the text embedding space using a trainable projection matrix. This projected text embedding is concatenated with a sequence of text embeddings that represent the user’s prompt. The combined sequence is then input LLM to generate the appropriate responses. To optimize performance, we propose a two-stage training paradigm for each component of the MLLM.
% \begin{figure}
%     \centering
%     \includegraphics[width=0.5\textwidth]{figures/vlm_structure.png}
%     \caption{Vision Language Model Deepfake Detection Pipeline (\xiangx{I think we can delete this figure.})}
%     \label{fig:vlm_structure}
% \end{figure}

\subsection{Automatic Image-Text Pair Data Generation} \label{sec:datagen}
\begin{figure}[t!]
    \centering
\includegraphics[width=\linewidth]{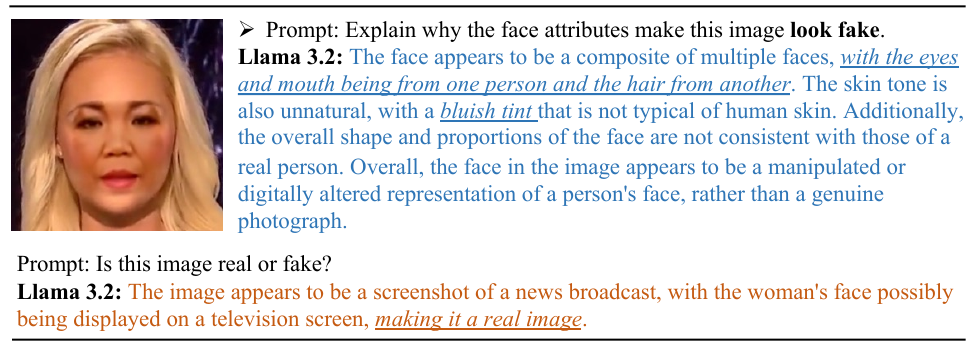}
    \caption{Automatic pseudo-text generation: comparing Llama 3.2's responses to prompts with and without labels.}
    \label{fig:data_gen1}
\end{figure}

Given the time-consuming nature of human annotation and the large size of typical deepfake datasets, we use a publicly available MLLM\footnote{In this paper, we use Llama 3.2~\cite{llama3.2}, but any publicly available generalist MLLM could be used in principle. } to scale the generation of pseudo-text pairs for all training images. We design a customized prompt that incorporates image labels as contextual information to prompt MLLM to generate the data: 
\textit{Explain why the face attributes (e.g., eyes, mouth, chin, hair, nose, and others) make this image look \textbf{Type}}. Here \textit{\textbf{Type}} denotes the binary (real/fake) image ground truth label. This label-based prompting method helps reducing hallucinations, ensuring that fake images consistently receive negative descriptions. The generated paragraph is then split into individual sentences based on keywords like ``mouth", ``eyes", and other facial landmarks. This added context helps the model generate more relevant and accurate pseudo-text pairs. %, thereby improving the model’s performance on deepfake detection tasks. 
Examples of the generated data are shown in \cref{fig:data_gen1}.

\begin{figure}[t!]
    \centering
    \includegraphics[width=\linewidth]{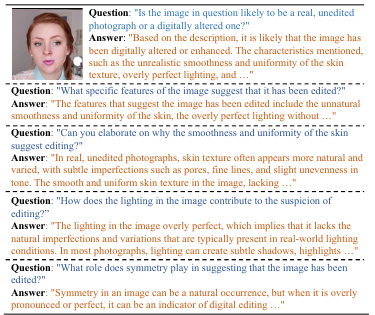}
    \caption{Example of instruction-tuning data generated from image captions (Sec. \ref{sec:datagen}).}
    \label{fig:instruct}
\end{figure}

To enhance multimodal alignment and deepfake reasoning in LLMs, we further generate instruction tuning data from image captions to teach the model to analyze visual inputs step by step, enabling reasoning for detecting deepfake artifacts \cite{llava-1.5,liu2023visual,vaillancourt2024instruction}. It also helps better alignment of image and language representations, ensuring structured and logical responses. We then use a LLM to generate diverse instructions and responses from image descriptions, creating a large dataset of image-centric, conversational training data for deepfake reasoning. We provide detailed examples of the generated instruction tuning data in \cref{fig:instruct}.

%\yifan{the subtitles should be consistent with the definition in the intro. E.g., this should be Alignment of Visual Embedding and Language Semantics}
\subsection{Aligning Deepfake Detection with Language} \label{align}
We design a specialized loss function to train our expert vision encoder. To enhance generalization, we incorporate image-text contrastive regularization to capture commonsense deepfake artifacts—features that can be described in natural language—alongside standard classification loss, which focuses on learning statistical deepfake artifacts. The overall loss function can be expressed as follows:
\begin{equation}
    \mathcal L_\mathrm{overall} = \mathcal L_\mathrm{cls}+ \mathcal L_\mathrm{cst}
    \label{eq:total}
\end{equation}
where $L_\mathrm{cls}$ is the the regular binary cross entropy loss and $L_\mathrm{cst}$ is the contrastive loss with details covered in \cref{uncertainty}.

A straightforward way to implementing this loss is to add a small adaptor that transforms the original vision embeddings for computing the classification loss while using the original vision embeddings for computing the contrastive loss. However, our experiments in \cref{tab:ablation} reveal that this design results in suboptimal performance. This occurs because the model overfits to easily learned statistical patterns from the classification loss, which hinders its ability to effectively capture the contrastive features. To address this, we propose an adaptive weighting mechanism that dynamically adjusts the contribution of each artifact type. For each image $x_i$, we branch its raw embedding from the vision encoder, denoted as $h_i$, into two sub-modules: one is used for contrastive learning to capture language-describable deepfake features, while the other passes through two self-attention layers to generate a transformed embedding that captures statistical deepfake features. We further introduce a trainable adapter, \( R^t \), that dynamically aggregates the original and transformed embeddings for each image, similar to the mixture-of-experts routing mechanism \cite{yu2024boosting}. For each image, denoting the contrastive feature as $z_{i}$ and the statistical feature as $v_i$, the aggregated embedding is computed as $e_{i} = w_{1} {v}_{i} + w_{2} z_{i}$ where \( \mathbf{w}_{i} = [w_{1}, w_{2}]\) represents the gating weights assigned by \( R^t \) to control the contribution of each representation. The gating weights are computed as $\mathbf{w}_{i} = \mathrm{Softmax}\big((R^t(v_{i}))\big)$ where \( R^t \) projects \( v_{i} \) to a 1-D vector. The elements of $\mathbf{w}_{i}$ sum to 1, representing the relative importance of each deepfake artifact. Finally, we apply a binary cross-entropy loss $\mathcal{L}_\mathrm{cls}$ on the \(e_{i}\) and the contrastive loss $\mathcal{L}_\mathrm{cst}$ on $z_i$. As shown in our ablation in \cref{tab:ablation}, this design yields 0.91\% AUC improvement on DFDC.

\subsection{Uncertainty Control for Text Noise} \label{uncertainty}
For contrastive learning, directly using the noisy pseudo-labels generated by a MLLM can introduce hallucinations and degrade detection performance. To tackle this, we draw inspiration from data uncertainty learning in face recognition \cite{chang2020data} and incorporate uncertainty control into contrastive learning \cite{clip}. Specifically, instead of using deterministic image embeddings, we adopt probabilistic embeddings \cite{shi2019probabilistic}, where each embedding is modeled as a distribution parameterized by a mean and variance. This allows for adaptive alignment and effectively mitigates the adverse effects of noisy pairs. To account for inconsistencies in image-text pairs, we model the raw vision embedding $h_{i}$ as a Gaussian distribution conditioned on the text description $T_i$: 
\begin{equation}
p(h_{i}|T_{i})=\mathcal{N}(h_{i}; \mu_{i}, \sigma_{i}^{2}I),
\end{equation}
where the mean and variance are predicted in an input-dependent manner by the backbone using a two-layer self-attention mechanism: $\mu_{i} = f_{\mu}(h_i), \  \sigma_i = f_{\sigma}(h_i)$, where $f_{(\cdot)}$ denotes the network parameters. Thus, the representation of each sample is no longer a deterministic embedding but a distribution. To allow the model to take gradients as usual, we apply the reparameterization trick \cite{kingma2013auto}. Specifically, we first sample random Gaussian noise \( \epsilon \sim \mathcal{N}(0, 1) \), and then generate the equivalent sampling representation as \( z_{i} = \mu_{i} + \sigma_i \cdot \epsilon \). Finally, a contrastive loss is applied to the resampled visual embeddings $z_{i}$ and textual embeddings $t_{i}=G(T_i)$ in a batch manner, where $G$ is the text encoder. This can be formally expressed as:
\begin{equation}
\mathcal{L}_{\text{cst}} = -\frac{1}{2B} \sum_{i=1}^{B} \left[ 
\log \frac{e^{w \cdot \tilde{z}_i \cdot \tilde{t}_i}}{\sum_{k=1}^{B} e^{w \cdot \tilde{z}_i \cdot \tilde{t}_k}} + 
\log \frac{e^{w \cdot \tilde{t}_i \cdot \tilde{z}_i}}{\sum_{k=1}^{B} e^{w \cdot \tilde{t}_i \cdot \tilde{z}_k}}
\right]
\end{equation}
\noindent In the equation, $\tilde{z}_i$, $\tilde{t}_i$ represent the normalized versions of $z_i$ and $t_i$, and $\omega$ is the temperature parameter.
%\eric{if use $p$ and $q$ in this formula, then let's try to be consistent before, avoid using $n$ as index. zhihua: p and q are index within a batch, n is on the entire dataset}

\subsection{Deepfake Reasoning with a LLM} \label{sec:llm}
To integrate deepfake detection with reasoning, we adopt a LLaVA-like \cite{llava-1.5} architecture for its superior accuracy in tasks like interpreting facial expressions and identifying object properties. We replace the original LLaVA vision encoder with our specialist vision encoder. Also, unlike the original implementation, which projects only image patch tokens to the LLM, we incorporate both a refined, discriminative class embedding and the original patch-wise tokens. This approach mitigates hallucinations and reduces inconsistencies between VQA detection and VQA reasoning, as the LLM is modulated with label information indicating whether the input image is real or fake. Formally, let \( f_{\text{proj}} \) denote this projection function:
\begin{equation}
e_{i}' = f_{\text{proj}}(v_{i}, e_{i}),
\end{equation}
where \( e_{i}' \in \mathbb{R}^{d_l} \) and \( d_l \) is the dimension of the language model's input. The projection \( f_{\text{proj}} \) aligns \( e_{i}' \) with the language model’s token space. Once aligned, these visual tokens, along with accompanying text tokens, are processed by a LLM (i.e., Vicuna \cite{vicuna2023}) to perform reasoning and generate responses based on both visual and textual inputs.

To train both the projection layer and the LLM, we employ instruction tuning using the auto-regressive loss. Given a triplet \((x, q, y)\) consisting of an image \(x\), question \(q\), and response \(y\) with \(L\) tokens, the model factorizes the probability of generating the sequence using the chain rule:
\begin{equation}
    p(y \mid x, q) = \prod_{i=1}^L p_{\theta}(y_i \mid x, q, y_{<i}),
    \label{eq:auto-regressive}
\end{equation}
Here, \(\theta=\{W, \phi\}\) represents the parameters of the projector and the LLM. The vision encoder weights are kept frozen.

\section{Experiments}
\label{sec:evaluation}

%\qin{I still find the results a little lacking. Would it be possible to have some robustness test and more visualizations with synthetic data with regional-manipulation? } 

\subsection{Experimental Settings}
\noindent \textbf{Datasets} \ We evaluate the effectiveness of our proposed framework in terms of generalizability and interpretability across multiple deepfake datasets:
% TODO change to the liner
\begin{enumerate} 
\item \textbf{FF++} dataset~\cite{ff++} consists of 1,000 real and 4,000 fake videos from various sources. Four deepfake techniques
%, namely DeepFakes (FF-DF)~\cite{deepfakes}, FaceSwap (FF-FS)~\cite{faceswap}, Face2Face (FF-F2F)~\cite{face2face}, and NeuralTextures (FF-NT)~\cite{neuraltexture}, 
are employed to generate the corresponding fake videos. We train all models over FF++ dataset. For evaluation, we adopt the c23 version of FF++.
\item \textbf{DFDC} dataset~\cite{dfdc} is a more challenging and larger deepfake detection dataset. Consistent with existing literature, we train the detector on the FF++ dataset and evaluate its performance on the DFDC dataset.
\item \textbf{DF40}~\cite{yan2024df40} is a recent dataset with 40 distinct forgery methods. We select 8 unseen faceswapping methods generated from the FF++ real samples for evaluation.
\item \textbf{DD-VQA} dataset~\cite{yan2024df40} is a recently established deepfake visual question answering dataset. It includes 14,782 question-answer pairs, with the images collected from the FF++  and human annotated text.
\end{enumerate}
%We follow the train/test split from the official GitHub repository, leading to a training set containing 114,885 images and 22,389 test images. For cross dataset evaluation, we utilize the c23 version of FF++ for training and other datasets for testing
%Besides the public dataset, we also conduct experiments on a self-collected deepfake instruction tuning dataset, DeepAnno. DeepAnno contains 33,875 and 1,324 images for the train/test sets, respectively. The real/fake images are collected from several publicly available face datasets, including DFDC~\cite{dfdc}, FF++~\cite{ff++}, ForgeryNet~\cite{forgerynet}, and ANTIFACT~\cite{artifact}, as well as synthesized by state-of-the-art generative models, such as StableDiffusion~\cite{sd}. Similar to DD-VQA, human inspectors are required to provide concrete text descriptions for each facial region regarding the authenticity. If the face region is considered to have artifacts, the description includes the specific reasons.

% \begin{figure}[h]
%     \centering
%     \includegraphics[width=0.45\textwidth]{figures/our_failure.png}
%     \caption{Failure cases of ours.}
%     \label{fig:our_failure}
% \end{figure}

\begin{table*}[]
\centering
\resizebox{\linewidth}{!}{
\begin{tabular}{l|c|ccccc|c|c}
\hline
\multirow{2}{*}{Method}                  & \multirow{2}{*}{Venues} & \multicolumn{5}{c|}{In-distribution test (AUC (\%)$\uparrow$)}                                 & Out-distribution test (AUC (\%)$\uparrow$) & Out-distribution test (ACC (\%)$\uparrow$) \\ \cline{3-9} 
                                         &                        & FF-DF          & FF-F2F         & FF-FS          & \multicolumn{1}{c|}{FF-NT}          & FF++  & DFDC                                       & DFDC                                       \\ \hline
UCF \cite{ucf}          & ICCV 2023              & 99.05          & \textbf{99.01} & \textbf{99.18} & \multicolumn{1}{c|}{95.27}          & 98.12 & 73.15                                      & 65.75                                      \\
SRM \cite{srm}          & CVPR 2021              & 97.82          & 97.08          & 97.17          & \multicolumn{1}{c|}{94.01}          & 96.52 & 68.44                                      & 62.83                                      \\
Face-X-Ray \cite{x-ray} & CVPR 2020              & 97.94          & 98.72          & 98.71          & \multicolumn{1}{c|}{92.90}          & 95.92 & 63.26                                      &   -                                         \\
SPSL \cite{spsl}        & CVPR 2021              & 97.81          & 97.54          & 98.29          & \multicolumn{1}{c|}{92.99}          & 96.10 & 70.40                                      &        62.35                                     \\
LSDA \cite{lsda}        & CVPR 2024              & 96.94              & 96.43              & 95.11              & \multicolumn{1}{c|}{94.92}              & 95.38     & 73.60                                      &    60.73                                      \\ \hline
\textbf{AuthGuard} (ours)                            & -                      & \textbf{99.65} & 98.83          & 98.85          & \multicolumn{1}{c|}{\textbf{98.13}} & \textbf{98.87} & \textbf{78.13}                             & \textbf{71.93}                                     \\ \hline
\end{tabular}
}
\caption{
Comparison on AUC (\%) is performed using a standard evaluation setup, where the models are trained on the FF++ dataset~\cite{ff++} and tested for in-distribution performance on FF++ and out-of-distribution performance on DFDC~\cite{dfdc}. %$^{*}$: Accuracy is based on LLM-generated 'real/fake' outputs, slightly surpassing the vision encoder's raw accuracy of 70.64\%. \yifan{R3's concern on if the gains we are seeing here are indeed coming from our claimed contribution is valid, can we add the ViT-backbone baseline and maybe call that PCL-ViT or something (assuming baseline is using PCL?)}
}
\label{tab:cls_acc}
\end{table*}

\begin{table*}[ht!]
\centering
\resizebox{\linewidth}{!}{
\begin{tabular}{l|c|ccccccc|c}
\hline
\multirow{2}{*}{Method}                  & \multirow{2}{*}{Venues} & \multicolumn{7}{c|}{Out-distribution test (AUC (\%) $\uparrow$)}                                                                            & Out-distribution test (ACC (\%) $\uparrow$) \\ \cline{3-10} 
                                         &                         & UniFace        & E4S            & FaceDancer     & FS-GAN         & InSwap         & \multicolumn{1}{c|}{SimSwap}        & \textbf{Average} & \textbf{Average}                            \\ \hline
RECCE \cite{cao2022end} & CVPR 2022               & 84.25          & 65.20          & 78.32          & 88.45          & 79.51          & \multicolumn{1}{c|}{73.04}          & 78.13            & 69.40                                           \\
CORE \cite{ni2022core}  & CVPR 2022               & 81.69          & 63.39          & 71.69          & 91.06          & 79.37          & \multicolumn{1}{c|}{69.34}          & 76.09            & 68.74                                           \\
SRM \cite{srm}          & CVPR 2021               & 78.24          & 66.73          & 77.43          & 84.52          & 76.15          & \multicolumn{1}{c|}{65.96}          & 74.84            & 68.11                                           \\
UCF \cite{ucf}          & ICCV 2023               & 78.67          & 69.17          & 80.06          & 88.09          & 76.85          & \multicolumn{1}{c|}{64.92}          & 76.29            & 68.26     \\                         
LSDA \cite{lsda}          & CVPR 2024               & 84.25          & 65.19          & 78.32          & 88.45          & 79.37          & \multicolumn{1}{c|}{69.34}          & 77.49            & 65.37                                           \\ \hline
\textbf{AuthGuard} (ours)            & -                       & \textbf{95.54} & \textbf{89.65} & \textbf{85.56} & \textbf{95.37} & \textbf{95.06} & \multicolumn{1}{c|}{\textbf{85.78}} & \textbf{91.16}   & \textbf{83.81}                                           \\ \hline
\end{tabular}
}
\caption{Accuracy comparison for several recent methods on six representative face-swapping techniques from the DF40 dataset~\cite{yan2024df40}. Our method outperforms the recent approaches by a substantial margin, leading to an 16.68\% improvement in performance. %$^{*}$: Accuracy is based on LLM-generated 'real/fake' outputs, slightly surpassing the vision encoder's raw accuracy of 82.96\%. \yifan{why only out-distribution test ACC has the LLM generated results-driven accuracy?}
}
\label{tab:df40}
\end{table*}

\noindent \textbf{Baselines} \
For the binary deepfake classification task, we compare our proposed framework with 5 state-of-the-art binary deepfake detectors, including UCF~\cite{ucf}, SRM~\cite{srm}, Face-X-Ray~\cite{x-ray}, SPSL~\cite{spsl}, and LSDA~\cite{lsda}. We use the implementation and pre-trained weights from the third-party evaluation toolbox, DeepfakeBench~\cite{deepfakebench}. For the deepfake reasoning task, we compare our method with BLIP-TI \cite{zhang2024common}, a state-of-the-art multi-modal VQA model specialized in deepfake detection and reasoning tasks.

\noindent \textbf{Metrics} \ For the binary deepfake detection task, following exsisting worsk \cite{ucf}, we report the AUC score on the FF++, DFDC, and DF40 datasets and compare them with state-of-the-art methods. To assess our model's reasoning capability against baselines on DD-VQA, we follow the methodology outlined in \cite{zhang2024common}. This includes using detection accuracy and four natural language processing metrics: BLUE-4\cite{bleu}, CIDEr~\cite{cider}, ROUGE\_L~\cite{rouge}, and METEOR~\cite{meteor}.

\noindent \textbf{Implementation Details} \ Our training process consists of two stages. In the first stage, we train an expert deepfake vision encoder using ViT-L/14~\cite{clip} for the vision branch and RoBERTa~\cite{roberta} for the text branch. Both ViT-L/14 and RoBERTa  are pretrained on the LAION dataset~\cite{laion}. During fine-tuning, the text encoder remains frozen, and only the vision encoder's weights are updated. The learning rate is set to 5e-6, with 1,000 warmup steps. We employ the Adam optimizer with a cosine learning rate scheduler and train for 5 epochs. The $\alpha$ and $\beta$ are set to 0.05 and 1 respectively. In the second stage, we perform instruction tuning for deepfake reasoning. We extract visual signals from the second-to-last layer of the vision encoder and use a two-layer MLP with GELU activation function as the vision language projector, while Vicuna-7B~\cite{vicuna} serves as the LLM. In the first sub-step, we fine-tune the weights in vision language projector 1 epoch and set the learning rate to 1e-3, the LLM and vision encoder are both frozen. In the second step, both projector and LLM are updated using LoRA~\cite{lora} for 1 epoch. The learning rate is set to 2e-5.

\begin{figure*}[ht!]
    \centering
    \includegraphics[width=1.0\textwidth]{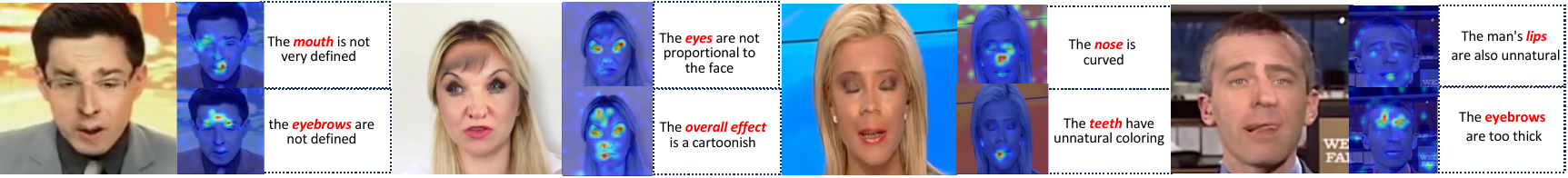}
    \caption{To qualitatively demonstrate the alignment between the vision representations from our expert vision encoder and the paired text, we visualize attention maps from our vision-language contrastive learning using the approach from \cite{chefer2021generic}. Text prompts are generated by LLAMA 3.2 \cite{llama3.2} with contextual labels, providing reasonable descriptions. For the same image, different descriptions of commonsense artifacts activate the corresponding facial regions, demonstrating that the representation learning is effectively guided by text.
    }
    \label{fig:attn}
\end{figure*}

\subsection{Evaluation Results}
\label{sec:detection_exp}
\noindent \textbf{Evaluation on Seen Attacks} We perform in-distribution evaluation with seen attacks using the FF++ dataset~\cite{ff++} and demonstrate that our vision encoder training by combining commonsense and statistical artifacts would not sacrifice the performance in this setting. As shown in \cref{tab:cls_acc}, when evaluating the in-distribution performance, our approach excels with an AUC of 98.87\% on the FF++ dataset~\cite{ff++}, outperforming the best baseline UCF (98.12\%)~\cite{ucf} by a margin of 0.76\%. 
When analyzing the performance on the individual subsets within FF++, our method exhibits comparable results to UCF on the FF-DF, FF-F2F, and FF-FS subsets. However, our approach significantly outperforms UCF on the FF-NT subset, achieving an AUC of 98.13\%, which exceeds UCF's 95.27\% by a substantial 3.00\%. These results show that our proposed vision encoder learning strategy maintains, if not improves, in-distribution performance compared to state-of-the-art methods.

\begin{table*}[h]
\centering
\resizebox{0.92\linewidth}{!}{
\begin{tabular}{l|cc|ccccc}
\toprule
\multirow{2}{*}{Method}& \multicolumn{2}{c|}{Deepfake Detection} & \multicolumn{5}{c}{Visual Question Answering (VQA)} \\ \cmidrule(lr){2-3} \cmidrule(lr){4-8}
& AUC (\%)$\uparrow$ & Accuracy (\%)$\uparrow$ & BLEU-4 $\uparrow$    & CIDEr $\uparrow$      & ROUGH\_L $\uparrow$  & METEOR $\uparrow$ & Average $\uparrow$\\
\midrule
% GPT4 \cite{gpt-4}                  & 0.70                    & \multicolumn{1}{c|}{--}     & \multicolumn{1}{c|}{--}     & \multicolumn{1}{c|}{--}     & --     \\

%InternVL2-8B \cite{chen2024internvl}                & - & 61.07                    & -     & -     & -     & -   & - \\
%Phi-3.5-Vision \cite{abdin2024phi}               & -    & 61.07                    & -  & -     & -    & -  & -  \\
General MLLMs (GPT-4o and \cite{abdin2024phi,chen2024internvl,li2024llava-onevision,gpt-4})  & -    & 58.43-75.00      & -  & -     & -    & -  & -  \\
BLIP-TI \cite{zhang2024common}           & -        & 87.49                    & 0.4075     & 2.0567     & 0.6085     & 0.3463  &  0.9823 \\
\midrule
LLaVA (off-the-shelf) \cite{llava-1.5}  & -   & 61.83                    & 0.1422     & 1.4231     & 0.2314     & 0.2489  & 0.5114 \\
LLaVA (fine-tuned)     & -   & 70.60                    & 0.3880     & 1.7320     & 0.5470     & 0.3212  &  0.7470  \\
\midrule
%LLaVA-DF (w/o semantic)    & 0.9011                    & \multicolumn{1}{c|}{x}     & \multicolumn{1}{c|}{x}     & \multicolumn{1}{c|}{x}     & x     \\
%LLaVA-DF (w/o class token) & 96.81 &90.11                         & 0.4850 & 3.2130& 0.6920        & \textbf{0.4120} &1.2005         \\
\textbf{AuthGuard} (ours)         &  \textbf{96.81} & \textbf{90.84}             & \textbf{0.4980} & \textbf{3.3050} & \textbf{0.6950} & 0.4010 & \textbf{1.2248}\\
\bottomrule
\end{tabular}
}
\caption{Comparison of various methods on the DD-VQA dataset~\cite{zhang2024common} for both deepfake detection and reasoning tasks. Our method achieved a 3.83\% improvement on deepfake detection and a 24.69\% improvement on the average quality metric on deepfake VQA.}
%\qin{Both the second last and the last rows are ours right? Do you want to include the second last row to highlight the importance of class token? If so, I feel the difference is quite minor and may not be a good datapoint. I would just remove it. }
\label{tab:ddvqa}
\end{table*}

\noindent \textbf{Evaluation on Unseen Attacks} \ We compare our method with the recent state-of-the-arts on out-of-distribution test using unseen attacks from the DFDC dataset \cite{dfdc} in \cref{tab:cls_acc} and selected unseen face-swapping attacks from DF40 \cite{yan2024df40} in \cref{tab:df40}.
In the out-of-distribution setting, our method achieves an AUC of 78.13\% on the DFDC dataset~\cite{dfdc} and 93.20\% on the DF40 dataset~\cite{yan2024df40}. 
These results represent significant improvements of 6.15\% and 16.68\%, respectively, compared to the best corresponding baselines \cite{ucf}. 
%This superior performance validates the effectiveness of using semantic information to regularize the visual representation learning process, leading to improved generalization capabilities for detecting unseen deepfake attacks. 
Based on the superior performance over state-of-the-art methods depicted in \cref{tab:cls_acc} and \cref{tab:df40}, we can conclude that our method surpasses existing methods not only in the in-distribution setting but also in the out-of-distribution setting. This validates the effectiveness of combining statistical and commonsense deepfake features, enhancing generalization and enabling robust detection of both seen and unseen deepfakes.

\begin{comment}
\begin{table}[t!]
\resizebox{\linewidth}{!}{
\begin{tabular}{l|c|c}
\toprule
MLLMs & Vision Encoder & Accuracy (\%)\\
\midrule
GPT4 \cite{gpt-4} & -- & 75.00 \\
\midrule
Llava-v1.5-7B \cite{llava-1.5} & \multirow{2}{*}{Clip-ViT-L/14-336px \cite{clip}} & 61.83\\
Phi-3.5-Vision \cite{abdin2024phi} &  & 61.07 \\
\midrule
InternVL2-8B \cite{chen2024internvl} & InternViT-6B-448px-V1-5 \cite{chen2024internvl} & 61.07\\
Llava-OV-7B \cite{li2024llava-onevision} & SigLip-ViT-L/14-384px \cite{zhai2023sigmoid} & 70.23 \\
\bottomrule
\end{tabular}
}
\caption{Comparing the accuracy of generic VLMs to detect Deepfakes on DDVQA dataset. \xiangx{Let's think where we can merge these results.}}
\label{tab:vlm_ff++}
\end{table}
\end{comment}

\begin{figure*}[ht!]
    \centering
\includegraphics[width=0.98\textwidth]{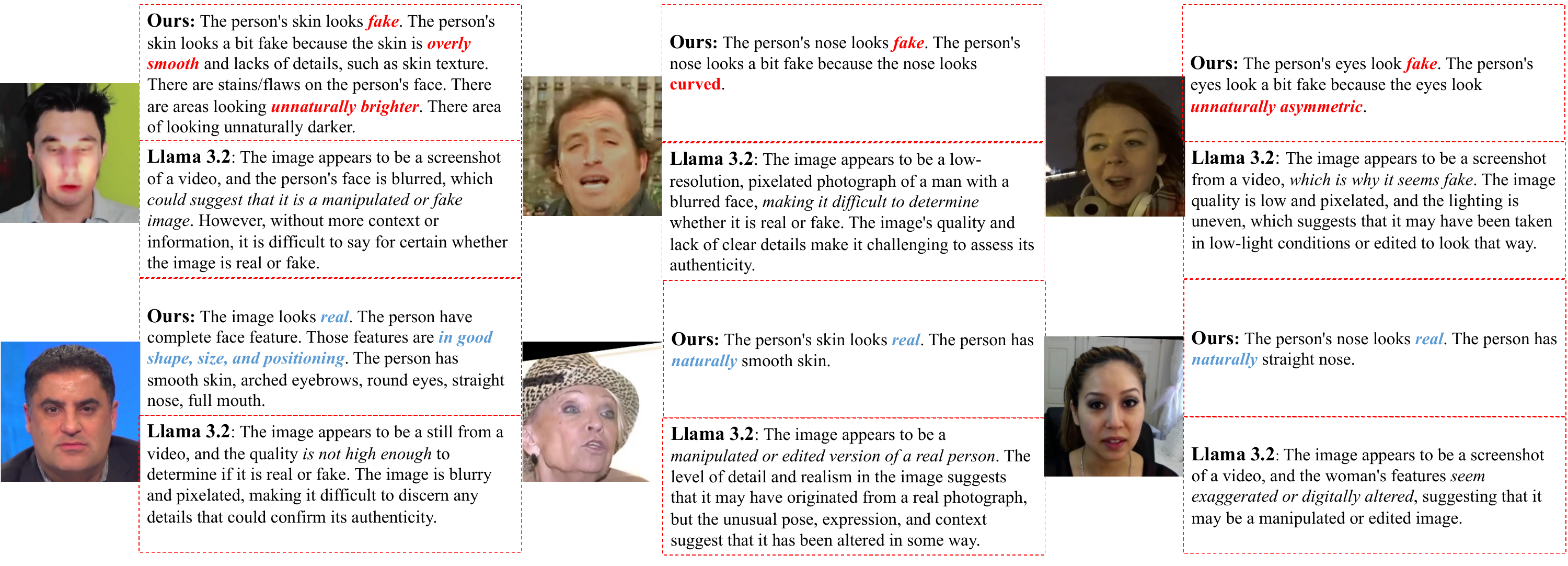}
    \caption{Reasoning examples of our method on DD-VQA. Highlighted text indicates the accurate descriptions of facial features.}
    \label{fig:ddvqa}
\end{figure*}

\noindent \textbf{Improving Explanation and Reasoning} \ 
In addition of detection accuracy, we evaluate deepfake explanation and reasoning capabilities of our method on the DD-VQA dataset~\cite{zhang2024common}, comparing it with general MLLMs and the expert MLLM, BLIP-TI~\cite{zhang2024common}, as shown in \cref{tab:ddvqa}.
Firstly, we benchmark the performance of GPT-4~\cite{gpt-4}, LLaVA-1.5-7B~\cite{llava-1.5}, InternVL~\cite{chen2024internvl}, Phi-3.5-Vision~\cite{abdin2024phi}, InternVL2-8B~\cite{chen2024internvl}, and LLaVA-OB-7B~\cite{li2024llava-onevision} for the deepfake detection task without any fine-tuning on the DD-VQA dataset. We prompt the MLLMs with ``\textit{Is this image real or fake?}" and evaluate accuracy by comparing the model's output (real/fake) with the ground truth labels. By this approach, these MLLMs obtain accuracy of 75.00\%, 61.83\%, 61.07\%, 61.07\%, and 70.23\%, respectively.
Additionally, we provide the detection and reasoning performance of the standard LLaVA model, trained on general vision data. 
While better than random guessing, its accuracy remains suboptimal, even for the supervised fine-tuned LLaVA on the same instruction tuning dataset. %This suggests that relying solely on a semantic MLLM model trained on general data fails to capture subtle facial anomalies, as some deepfake-specific artifacts are non-describable and not well-represented by the vision encoder. 
These limitations in standard MLLMs highlight the need to enhance the vision encoder with deepfake-specific knowledge. On the DD-VQA dataset, BLIP-TI~\cite{zhang2024common} is the only deepfake domain expert model providing both capabilities on binary detection (yes/no) and reason explanation. 
Compared to BLIP-TI~\cite{zhang2024common}, our method can produce probabilistic output and achieve a detection accuracy of 90.84\%, outperforming BLIP-TI by 3.83\% on deepfake detection accuracy. This improvement in detection accuracy validates the effectiveness of our approach in accurately identifying deepfake artifacts within the DD-VQA dataset~\cite{zhang2024common}.
In reasoning, our method significantly outperforms BLIP-TI across all four metrics and 24.69\% improvement on average. 
Specifically, we achieve scores of 0.4980 for BLEU-4, 3.3050 for CIDEr, 0.6950 for ROUGE\_L, and 0.4010 for METEOR, while BLIP-TI scores 0.4075, 2.0567, 0.6085, and 0.3463, respectively. These higher scores indicate that the responses generated by our method are better aligned with human annotators' assessments, and that the binary token produced by our method shows greater consistency with the labels.

\subsection{Ablation Study and Visualization}
\label{exp:ablation}

\begin{table}[]
\centering
\resizebox{0.9\columnwidth}{!}{%
\begin{tabular}{ccc|cc}
\toprule
\multicolumn{3}{c}{Modules} & \multicolumn{2}{c}{Datasets} \\ \cmidrule(lr){1-3} \cmidrule(lr){4-5}
Semantic Artifacts & Uncertainty Est. & Adapter & FF++ & DFDC \\
\midrule
     &  &                           & 98.35         & 75.22         \\
\checkmark &   &                     & 98.69         & 76.29         \\
\checkmark & \checkmark  &                     & 98.62         & 77.20         \\
\checkmark & \checkmark  &  \checkmark  & \textbf{98.87} & \textbf{78.13} \\ 
\bottomrule
\end{tabular}%
}
\caption{Ablation study on AUC (\%), evaluating the improvement of the proposed modules on the FF++ and DFDC datasets.} %\yifan{seems that adding probablistic embedding made the accuracy worse on FF++? given this and the fact that we are just using standard probabilistic embedding from the FR literature, do we still want to put it as a separate section in methodology and have an ablation on it? what about tone it down and just absorb it into alignment of visual embedding and language semantics (semantic learning).}
\label{tab:ablation}
\end{table}

\begin{figure*}[t!]
    \centering
    \includegraphics[width=\linewidth]{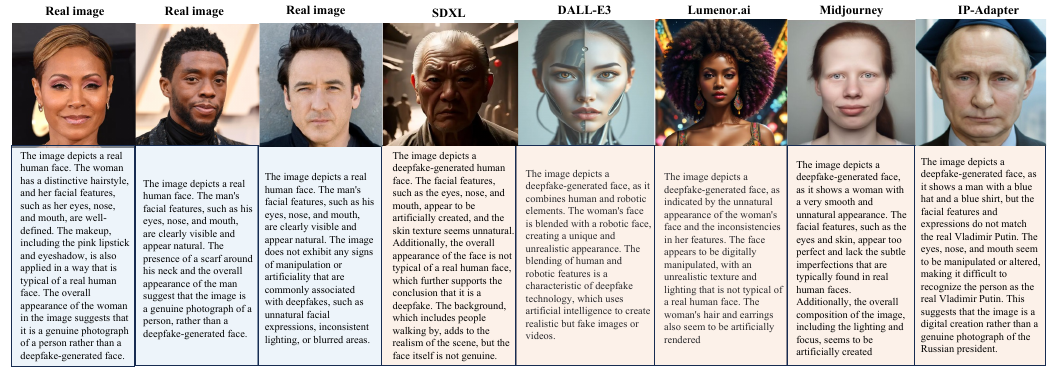}
    \caption{Evaluating deepfake images beyond FF++: the images with blue-shaded texts are real; the remaining images are AI-generated.}
    \label{fig:more-vis}
\end{figure*}

\noindent\textbf{Impact of Various Image Encoder Improvements} \ To evaluate the impact of each proposed module, we performed ablation studies by incrementally adding each component to the image encoder training process depicted in \cref{tab:ablation}. First, incorporating semantic learning through image-text contrastive learning led to an AUC increase from 98.35\% to 98.69\% on FF++ and from 75.22\% to 76.29\% on DFDC, suggesting that semantic alignment enhances performance even in the presence of noisy labels. Next, we introduced an uncertainty estimation (Uncertainty Est.) module to address the unreliability of pseudo-text annotations, further improving the AUC from 76.29\% to 77.20\%. This result indicates that probabilistic embeddings can strengthen the effectiveness of semantic learning by providing adaptive vision-language alignment. Finally, with the integration of adaptive balancing between commonsense and forgery-specific features, the AUC rose from 75.22\% to 78.13\%, achieving the highest scores across all configurations. These findings demonstrate that both commonsense and statistical artifacts are crucial for deepfake detection, and adaptively combining these artifacts offers an optimal solution.

\noindent\textbf{Interpretation and Reasoning} \ 
%For reasoning training, we incrementally adapt the off-the-shelf LLaVA model to our specific requirements, assessing its performance on the DD-VQA dataset. The results, shown in Table~\ref{tab:ddvqa}, reveal that even with the default fine-tuning approach—where the vision encoder is fixed and only the vision-language projector and LLM are tuned—the model achieves a detection accuracy of just 0.7060 and a BLEU-4 score of 0.3880. This emphasizes the limitations of using a vision encoder pre-trained on natural images, supporting our hypothesis that such encoders are insufficient for extracting deepfake-relevant features.
%In the fifth row, after replacing the standard vision encoder with our specialized deepfake detection encoder, both detection and reasoning performance significantly improve, reaching an accuracy of 0.9084 and a BLEU-4 score of 0.4980. This substantial improvements demonstrates the efficacy of our expert vision encoder in capturing deepfake-specific artifacts for more accurate detection and reasoning.
For reasoning training, we incrementally adapt the off-the-shelf LLaVA model to our specific requirements, assessing its performance on the DD-VQA dataset \cite{zhang2024common}. \cref{tab:ddvqa} reveal that with the default fine-tuning approach where the CLIP \cite{clip} vision encoder is fixed and only the vision-language projector and LLM are tuned, the model achieves a detection accuracy of just 70.60\% and a BLEU-4 score of 0.3880. This shows limitations of using a vision encoder pre-trained on natural images, supporting our hypothesis that such encoders are insufficient for extracting deepfake-relevant features.
To address this issue, we fine-tune the standard vision encoder with our specialized deepfake detection encoder, which is trained to capture both commonsense and statistical artifacts. After this modification, we observe significant improvements in both detection and reasoning performance. 
The model achieves an accuracy of 90.84\% for deepfake detection and a BLEU-4 score of 0.4980 for reasoning, outperforming the baseline LLaVA model by a substantial margin.
These results demonstrate the efficacy of our expert vision encoder in capturing deepfake-specific artifacts, enabling more accurate detection and improved reasoning capabilities. %By incorporating both semantic and forgery-specific information during the encoder's training process, our approach addresses the limitations of relying solely on pre-trained vision models designed for natural images.
%The substantial improvements achieved by our method highlight the importance of specialized encoders tailored for the unique challenges posed by deepfake detection.
%\xiangx{Note this paragraph can be merged with subsection: improving explanation and reasoning} \yifan{there are two factors conflating the conclusion here. In the beginning you talk about fine-tuning strategy of MLLM, later you talk about replacing standard vision encoder in Llava with ours. Which point are you trying to highlight? for the former, I expect results of different fine-tuning strategies (just train the connector, pretrain connector with LLM frozen and then fine-tune both connector and LLM etc.)}
%Additionally, our proposed inference calibration strategy further enhances detection accuracy to 0.9160. Figure~\ref{fig:calibrate} illustrates how calibration resolves conflicts and improves the VLM performance. In the first round of evaluation, the VLM fails to identify the test image as a deepfake, resulting in an inaccurate description. However, the logits from the vision encoder suggest that the image is likely artificial. In the second round, the VLM is provided with these classification logits and asked to reassess the image. This re-evaluation process calibrates the VLM's prediction, leading it to successfully generate the correct classification and identify problematic face regions.

%\subsection{Visualization} 
%In addition to overall metrics, we provide concrete examples to showcase the interpretability of our proposed method from various angles. 
\noindent\textbf{Attention Map Visualization} \ 
%To further demonstrate the effectiveness of our semantic learning module, we visualize the attention maps \cite{chefer2021generic} produced by our model in Figure~\ref{fig:attn}. The text prompts shown are pseudo-labels generated by LLAMA3.2, and all four images are fake.As illustrated, using different text prompts on the same deepfake image successfully guides the encoder to focus on the targeted regions, which aligns with our objective of encouraging the model to learn semantic artifacts directed by pseudo-text. Observing Fig.~\ref{fig:attn}, it is clear that the attention masks correspond well with the paired text descriptions, indicating that our vision encoder effectively captures the relevant semantic artifacts as guided by the text.
To further validate the efficacy of our proposed semantic learning module, we present a visual analysis of the attention maps~\cite{chefer2021generic} generated by our model, as depicted in \cref{fig:attn}. The text prompts displayed are pseudo-labels derived from the Llama 3.2 model \cite{llama3.2}, and all four images utilized in this analysis are synthetic deepfakes.
As illustrated, by providing distinct text prompts for the same deepfake image, our model successfully guides the encoder to concentrate on the targeted regions, aligning with our objective of encouraging the model to learn semantic artifacts directed by pseudo-text descriptions. An examination of \cref{fig:attn} reveals a strong correspondence between the attention masks and the associated text descriptions, indicating that our vision encoder effectively captures the relevant semantic artifacts as guided by the textual input.

\noindent\textbf{Qualitative Examples of Deepfake Explanation} \ 
%Fig~\ref{fig:ddvqa} illustrates four deepfake images from DD-VQA along with the corresponding answers generated by our model. Our model successfully identifies all four samples as deepfakes and offers detailed explanations for each. For the first image, the model highlights the overly smooth skin and misaligned mouth area as indicators of fakery. In the second image, it points out the unnaturally bright facial area. For the third and fourth examples, the model notes suspicious features in the hair and eyebrows, aligning with human perception. These detailed explanations enhance the interpretability of our deepfake detector beyond simple binary results. 
\cref{fig:ddvqa} presents qualitative results on the DD-VQA dataset \cite{zhang2024common}, comparing our model's responses with Llama 3.2. Our model accurately classifies all samples (top three: deepfake, bottom three: real) with precise explanations. %Specifically, it identifies unnaturally bright facial regions in the first example and detects structural anomalies in the second case, where the nose appears curved and misaligned with the face structure.
While Llama 3.2 \cite{llama3.2} often expresses uncertainty in detection and provides only high-level explanations, our expert encoder-enhanced approach delivers both accurate classification and detailed semantic reasoning, particularly in identifying specific facial artifacts and structural inconsistencies commonly found in deepfakes. To evaluate our model’s generalization, we test its performance on facial images from diverse generation methods (e.g., SDXL \cite{podell2023sdxl} and IP-Adapter \cite{ye2023ip}). As shown in \cref{fig:more-vis}, our model provides reasonable answers to queries, demonstrating its effectiveness beyond FF++ and its adaptability to recent image synthesis methods. %This highlights the advantages of our generated instruction data, which improves the descriptiveness of semantically meaningful anomalies in synthetic images and enhances reasoning capabilities.
\section{Conclusion}
This paper presents \textbf{AuthGuard}, a unified deepfake detection and reasoning framework that boosts both the generalization and interpretability of deepfake detection by combining statistical and commonsense deepfake artifacts. 
%The framework features three key components: 1) a deepfake-specialist vision encoder trained with contrastive learning on pseudo-labeled data, 2) uncertainty learning to handle label noise in the pseudo-labeled data, and 3) a large language model to support downstream reasoning. 
Extensive evaluations show that our approach outperforms existing methods in detection accuracy, generalization (6.15\%), and interpretability (24.69\%). By bridging specialized deepfake detection with multi-modal large language models, we take a step toward more transparent and generalizable deepfake detection systems. We hope our work contributes to combating misinformation, protecting digital identity, and fostering trust in media authenticity.
%This paper introduces a method\qin{unified framework?} to enhance deepfake detectors' generalizability and interpretability through fine-grained language supervision\qin{I think this is not accurate, our method is mostly about leveraging a deepfake-expert language-supervised vision encoder to boost LLAVA reasoning performance for deepfake. You may want to be more specific and accuarate in conclusion.}. Our approach involves two stages: First, we develop a deepfake vision transformer encoder using CLIP-like contrastive learning, balancing contrastive and discriminative objectives to improve generalization and prevent overfitting. In the second stage, we integrate this encoder with a large language model to create a vision-language model for deepfake image interpretation. We refine the model with custom deepfake vision instruction tuning data for better feature alignment. Evaluations on multiple datasets show that our method outperforms existing detectors in both accuracy and interpretability. 
%\yifan{we should also add a section on limitations, it can be clarify of assumptions that we make and (optionally) implications if these assumptions do not hold. E.g., one limitation is that we focus on semantic artifacts that can be described by language while we acknowledge that there can be semantics that can't be linguistically described. We to future works on modeling of such non language describable artifacts.}

% \newpage

\small
\bibliographystyle{ieeenat_fullname}
% \bibliography{main}

% \twocolumn
% {
%     \small
%     \bibliographystyle{ieeenat_fullname}
%     \bibliography{main}
% }
\end{document}